\documentclass[conference]{IEEEtran}
\IEEEoverridecommandlockouts

\usepackage{cite}
\usepackage{amsmath,amssymb,amsfonts}
\usepackage{algorithmic}
\usepackage{graphicx}
\usepackage{textcomp}
\usepackage{xcolor}

\usepackage{algorithm}
\usepackage{booktabs}
\usepackage{array}
\usepackage{tgpagella}
\usepackage{multirow} 
\usepackage{colortbl}        
\usepackage{caption}         
\usepackage{threeparttable}  
\usepackage{amssymb}

\usepackage[caption=false,font=normalsize,labelfont=sf,textfont=sf]{subfig}
\usepackage{stfloats}
\usepackage{url}
\usepackage{verbatim}
\usepackage{titlesec}
\usepackage{hyperref}
\hypersetup{
  colorlinks=true,
  linkcolor=blue,
  urlcolor=blue,
  citecolor=blue,
}
\usepackage{balance}

\def\BibTeX{{\rm B\kern-.05em{\sc i\kern-.025em b}\kern-.08em
    T\kern-.1667em\lower.7ex\hbox{E}\kern-.125emX}}
\begin{document}

\title{Explainability-Guided Defense: Attribution-Aware Model Refinement Against Adversarial Data Attacks\\
}

\author{\IEEEauthorblockN{Longwei Wang$^\dagger$, Mohammad Navid Nayyem$^\dagger$, Abdullah Al Rakin$^\dagger$, KC Santosh$^\dagger$, Chaowei Zhang$^\ddagger$, \\ and Yang Zhou$^\#$}
\IEEEauthorblockA{\textit{$^\dagger$AI Lab, Department of Computer Science}, University of South Dakota, Vermillion, SD, USA\\
$^\ddagger$\textit{School of Information Engineering}, Yangzhou University, Yangzhou, Jiangsu, China\\
\textit{$^\#$Department of CSSE}, Auburn University, Auburn, AL, USA\\
\{longwei.wang $|$ kc.santosh\}@usd.edu, cwzhang@yzu.edu.cn, and yangzhou@auburn.edu
}
}


\maketitle

\begin{abstract}

The growing reliance on deep learning models in safety-critical domains such as healthcare and autonomous navigation underscores the need for defenses that are both robust to adversarial perturbations and transparent in their decision-making. In this paper, we identify a connection between interpretability and robustness that can be directly leveraged during training. Specifically, we observe that spurious, unstable, or semantically irrelevant features identified through Local Interpretable Model-Agnostic Explanations (LIME) contribute disproportionately to adversarial vulnerability. Building on this insight, we introduce an attribution-guided refinement framework that transforms LIME from a passive diagnostic into an active training signal. Our method systematically suppresses spurious features using feature masking, sensitivity-aware regularization, and adversarial augmentation in a closed-loop refinement pipeline. This approach does not require additional datasets or model architectures and integrates seamlessly into standard adversarial training. Theoretically, we derive an attribution-aware lower bound on adversarial distortion that formalizes the link between explanation alignment and robustness. Empirical evaluations on CIFAR-10, CIFAR-10-C, and CIFAR-100 demonstrate substantial improvements in adversarial robustness and out-of-distribution generalization.

\end{abstract}


\section{Introduction}
Deep neural networks (DNNs) have revolutionized modern artificial intelligence, achieving state-of-the-art performance across a wide spectrum of applications, including image recognition, natural language processing, autonomous navigation, and medical diagnostics~\cite{krizhevsky2012imagenet, he2016deep}. Among these, convolutional neural networks (CNNs) have emerged as the cornerstone for visual perception tasks, owing to their ability to hierarchically extract and integrate spatial patterns through shared-weight architectures and local connectivity.
Despite their remarkable capabilities, DNNs remain acutely vulnerable in adversarial and out-of-distribution (OOD) environments. A critical limitation is their susceptibility to adversarial examples—imperceptible perturbations carefully optimized to mislead models into making incorrect and often high-confidence predictions~\cite{goodfellow2015explaining, madry2018towards}. These perturbations exploit the model's reliance on brittle and non-semantic features, reflecting the underlying instability of decision boundaries in high-dimensional spaces. Beyond adversarial attacks, recent studies have shown that DNNs frequently latch onto spurious correlations or shortcut cues—non-causal features that correlate with labels in the training data but fail under distributional shifts~\cite{geirhos2019shortcut}. Such dependencies can severely undermine a model’s generalization and fairness, especially in safety-critical applications like healthcare and autonomous driving.
Compounding these challenges is the opaque nature of modern deep learning models. Often characterized as ``black boxes,'' these models offer limited insight into the internal logic behind their predictions~\cite{samek2017explainable}. This opacity hampers model debugging, restricts regulatory compliance, and diminishes user trust, thereby constraining the deployment of AI in domains that demand accountability, transparency, and robustness.

To address these issues, explainable AI (XAI) methods such as Local Interpretable Model-Agnostic Explanations (LIME)~\cite{ribeiro2016why} and SHapley Additive exPlanations (SHAP)~\cite{lundberg2017unified} have emerged as powerful tools for attributing predictions to input features. These techniques expose model reasoning by generating instance-specific feature importance maps, which can reveal reliance on irrelevant or misleading patterns. However, most XAI methods are used retrospectively—they analyze trained models without influencing the training process itself. As a result, a critical opportunity remains untapped: using attribution feedback to actively guide learning away from spurious features during model optimization.

\begin{figure}[tbp]
  \centering
  \includegraphics[width=0.99\linewidth]{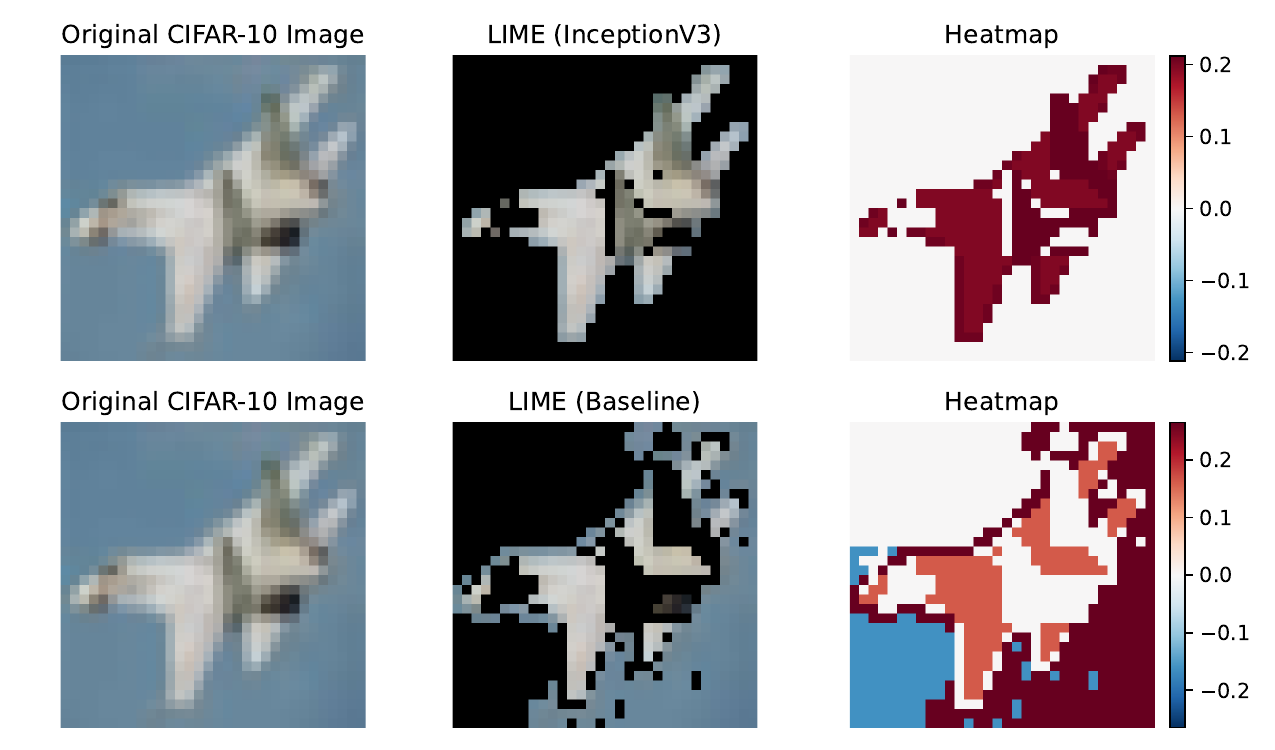}
  \caption{LIME-based feature attribution maps illustrating the difference in learned representations between robust and non-robust models. The robust model (InceptionV3) consistently attends to semantically meaningful regions, while the non-robust baseline focuses on irrelevant or noise-sensitive areas.}
  \label{fig:heatcomp}
\end{figure}

While adversarial training is a widely adopted defense mechanism, it often acts as a brute-force solution, lacking interpretability and failing to provide insights into how and why robustness is achieved. Recent empirical findings, however, suggest a deeper connection: robust models tend to produce more stable and semantically aligned explanations, while non-robust models exhibit scattered or spurious attribution patterns~\cite{ross2018improving, dombrowski2019explanations}. This relationship is visually demonstrated in \textbf{Figure~\ref{fig:heatcomp}}, where robust models consistently attend to task-relevant features. These insights motivate a paradigm shift—moving beyond reactive robustness toward proactive, explanation-aware training.

In this paper, we propose a novel training framework that operationalizes this insight by embedding explainability directly into the adversarial learning pipeline. Specifically, we introduce an attribution-aware refinement architecture that transforms LIME from a post-hoc explanation tool into a functional training signal. Our framework identifies semantically irrelevant, overly sensitive, or unstable features using LIME attributions and systematically suppresses their influence through a triad of mechanisms: (1) input-level feature masking, (2) sensitivity-aware gradient regularization, and (3) targeted adversarial training. This closed-loop architecture iteratively recalibrates the model’s attention, aligning learned representations with robust, semantically grounded features.
Our central hypothesis is that interpretability and robustness are mutually reinforcing. Models that ground predictions in interpretable and stable features are less susceptible to adversarial manipulation, while robust models in turn produce more faithful and reliable explanations. We formalize this intuition by deriving an attribution-aware lower bound on adversarial distortion using a masked gradient formulation inspired by local Lipschitz continuity. Empirically, we show that our framework enhances both robustness and attributional clarity across a range of evaluation settings. The key contributions of this work are summarized below:
\begin{itemize}
    \item We propose a unified defense framework that transforms interpretability into an active training signal, systematically reducing spurious feature dependencies via LIME-guided model refinement.
    \item We introduce a theoretical robustness analysis based on attribution-aligned gradient suppression and local Lipschitz bounds, establishing formal links between interpretability and adversarial resilience.
    \item We conduct extensive experiments on CIFAR-10, CIFAR-10-C, and CIFAR-100 demonstrating substantial improvements in adversarial robustness and attributional quality without compromising clean accuracy.
\end{itemize}

\section{Related Works}


\subsection{Adversarial Robustness in Deep Models}
Convolutional neural networks (CNNs) are known to be vulnerable to adversarial perturbations—carefully crafted, imperceptible changes to input data that can mislead the model into making incorrect predictions~\cite{szegedy2014intriguing,zhu2024neural}. Foundational attacks such as the Fast Gradient Sign Method (FGSM)~\cite{goodfellow2015explaining} and Projected Gradient Descent (PGD)~\cite{madry2018towards} have prompted the development of a variety of defenses, most notably adversarial training~\cite{madry2018towards,zou2024improving,wang2025bridging,wang2024enhanced} and preprocessing-based sanitization methods~\cite{qin2024apbench}.

Although effective, these approaches often incur substantial computational costs and may generalize poorly across different attack types~\cite{ranabhat2025multi,wang2024dense,wang2024enhanced}. More importantly, they are typically agnostic to the model’s internal reasoning and provide little insight into whether the learned representations are semantically meaningful or aligned with human intuition.

\subsection{Model Interpretability}

Interpretability methods aim to provide insight into the decision-making processes of complex models. Popular tools such as Local Interpretable Model-Agnostic Explanations (LIME)~\cite{ribeiro2016why} and SHapley Additive exPlanations (SHAP)~\cite{lundberg2017unified} attribute importance scores to input features based on their contribution to the model’s output. Visual explanation techniques such as Grad-CAM~\cite{selvaraju2017grad} and Integrated Gradients~\cite{sundararajan2017axiomatic} have further enhanced the interpretability of CNNs, particularly in vision tasks.

However, most of these techniques are employed post hoc and play no role in the training process \cite{wang2024explainable}\cite{ennab2025advancing,wang2019representation,wall2025winsor,uddin2025expert}. Consequently, while they are useful for diagnosing spurious behavior or identifying misleading dependencies, they do not directly contribute to improving model robustness. Our work departs from this passive paradigm by actively integrating LIME into the training loop to guide attention control and improve robustness.

\subsection{Integrating Interpretability and Robustness}

There is growing recognition that interpretability can play a pivotal role in achieving robust deep learning \cite{chander2025toward}\cite{seth2025bridging}\cite{wang2021explaining,jagatheesaperumal2024enabling}\cite{wang2025explainability,nayyem2024bridging}. Doshi-Velez and Kim~\cite{doshi2017rigorous} suggested that aligning models with human-interpretable features may help avoid reliance on spurious patterns. Adebayo et al.~\cite{adebayo2018sanity} proposed sanity checks to evaluate the reliability of saliency maps, while Slack et al.~\cite{slack2020fooling} showed that explanation tools like LIME and SHAP are themselves vulnerable to adversarial manipulation.

Ross and Doshi-Velez~\cite{ross2018improving} demonstrated that incorporating gradient-based interpretability constraints into the loss function can improve adversarial robustness. Similarly, Dombrowski et al.~\cite{zhou2022feature} studied how adversarial robustness affects the stability of feature attribution maps. While these studies highlight a conceptual link between robustness and interpretability\cite{csahin2025unlocking}\cite{zhou2025advancing}, they stop short of proposing a unified training framework that exploits this relationship for model refinement.



\section{Methodology}

\begin{figure}[tbp]
    \centering
    \includegraphics[width=0.99\linewidth]{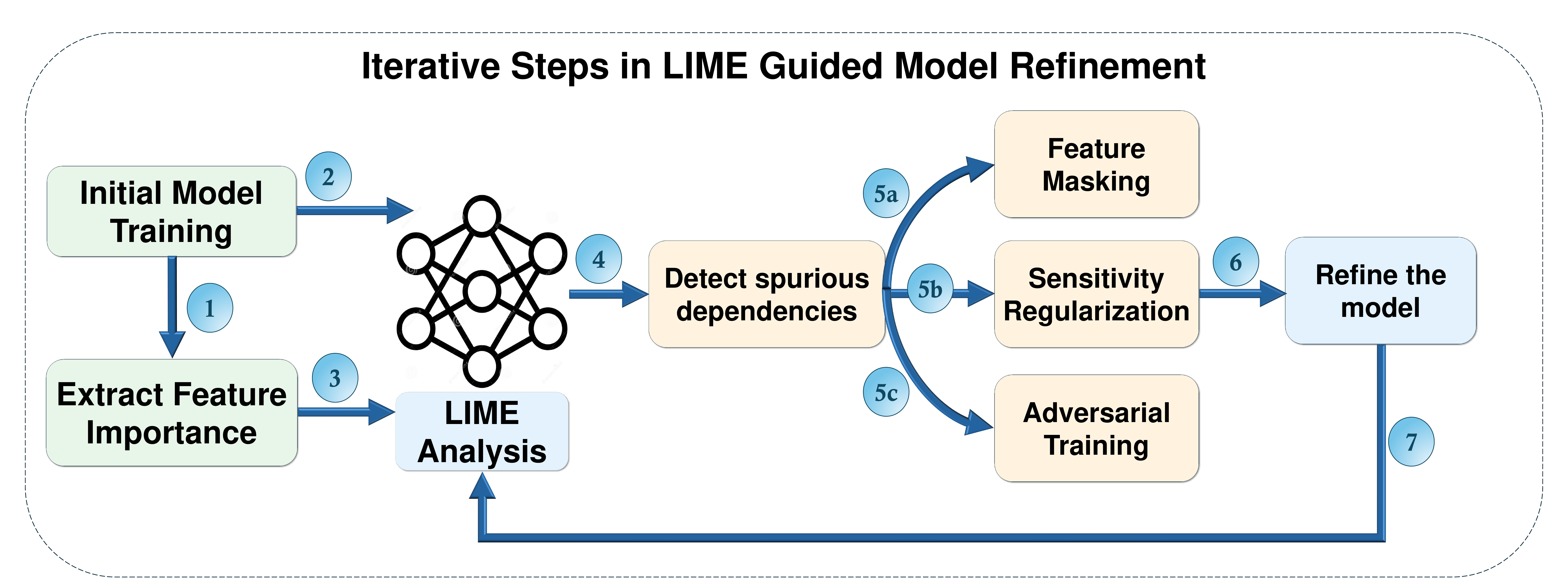}
    \caption{Overview of the iterative XAI-guided refinement framework. LIME highlights spurious features post initial training. These are mitigated through feature masking, sensitivity regularization, and adversarial training. The process iterates until robustness and interpretability metrics converge.}
    \label{fig:lime_steps}
\end{figure}

To bridge interpretability and robustness, we introduce a unified framework that integrates explainable AI (XAI) into the training loop of deep neural networks for targeted model refinement under adversarial conditions. At the core of our approach is the use of Local Interpretable Model-Agnostic Explanations (LIME) to identify spurious, unstable, or semantically irrelevant features that compromise robustness. Rather than using interpretability tools as passive diagnostics, we transform them into active refinement signals that guide an iterative training process involving feature masking, sensitivity regularization, and adversarial augmentation. As shown in Fig.~\ref{fig:lime_steps}, the process is cyclic and designed to promote semantic consistency, attributional stability, and gradient robustness over successive refinements.

\subsection{Attribution-Guided Analysis}
To assess feature-level relevance, we apply LIME to construct local surrogate models around each input \( x \). Let \( \mathcal{Z} = \{(z_i, f(z_i))\}_{i=1}^N \) be a set of perturbed inputs obtained by masking subsets of \( x \), and let:
\begin{equation}
w_i = \exp\left(-\frac{\|x - z_i\|_2^2}{\sigma^2}\right)
\end{equation}
be a locality-aware kernel. A linear surrogate model \( g(z) = \beta_0 + \sum_{j=1}^d \beta_j z_j \) is trained by minimizing the weighted squared loss:
\begin{equation}
\mathcal{L}_{\text{surrogate}} = \sum_{i=1}^N w_i \left( f(z_i) - g(z_i) \right)^2.
\end{equation}
The resulting attribution vector \( \beta \) serves as a feature-level explanation of the model’s decision. The absolute attribution \( |\beta_j| \) quantifies the importance of input feature \( x_j \):
\begin{equation}
\text{Importance}(x_j) = |\beta_j|.
\end{equation}







\subsection{Spurious Feature Identification}

Not all highly attributed features are beneficial for robustness. We define features as \textit{spurious} if they exhibit at least one of the following undesirable traits:

\textbf{(1) Irrelevance:}  
A feature is deemed irrelevant if it receives disproportionately high attribution from LIME despite lacking semantic relevance. 
To distinguish relevant from irrelevant attributions, we leverage a reference model with higher interpretability—specifically, a well-trained ResNet-50—as a proxy for identifying class-relevant features. We then compare the LIME attribution maps of our baseline model against those of the reference model to assess alignment with the true class semantics. A feature is flagged as irrelevant if it exceeds a predefined attribution magnitude threshold:
\begin{equation}
|\beta_j| > \tau,
\end{equation}
where \( \beta_j \) denotes the LIME attribution score for feature \( x_j \). Features satisfying this condition but not supported by the reference model are considered spurious and targeted for suppression in the refinement process.

\textbf{(2) Sensitivity:}  
Features with excessively high input gradients can cause disproportionate changes in the model’s output under small perturbations. To quantify this behavior, we define the sensitivity of a feature \( x_j \) as the magnitude of its input gradient:
\begin{equation}
\text{Sensitivity}(x_j) = \left| \frac{\partial f(x)}{\partial x_j} \right| > \epsilon,
\end{equation}
where \( \epsilon \) is a predefined sensitivity threshold. To ensure that gradient magnitude aligns with semantic relevance, we additionally compare feature sensitivities with those from a reference model (e.g., a well-trained ResNet-50) known for producing more interpretable attributions. If a feature exhibits high sensitivity in the baseline model but lacks corresponding importance in the reference model, it is considered semantically ungrounded and is subsequently targeted for regularization during the model refinement process.

\textbf{(3) Instability:}  
Features whose attributions vary significantly under input perturbations are considered unstable. Attribution instability is defined as the variance of LIME importance scores across \( N \) perturbed samples:
\begin{equation}
\text{Instability}(x_j) = \text{Var} \left( \left\{ \beta_j^{(i)} \right\}_{i=1}^N \right) > \delta,
\end{equation}
where \( \delta \) is an instability threshold. A feature is classified as spurious if it satisfies at least one of the above conditions:
\begin{align}
x_j \in \mathcal{F}_{\text{spurious}} \iff \ & |\beta_j| > \tau \quad \lor \quad \text{Sensitivity}(x_j) > \epsilon \notag \\
& \lor \quad \text{Instability}(x_j) > \delta.
\end{align}

\subsection{XAI-Guided Model Refinement}

\subsubsection{Feature Masking}
We suppress the influence of spurious features using a binary mask \( m \in \{0, 1\}^d \), where:
\begin{equation}
m_j = \begin{cases}
0 & \text{if } x_j \in \mathcal{F}_{\text{spurious}}, \\
1 & \text{otherwise}.
\end{cases}
\end{equation}
The masked input is then
\begin{equation}
x^{\text{masked}} = x \odot m.
\end{equation}

\subsubsection{Sensitivity Regularization}
To penalize over-reliance on unstable features, we introduce a sensitivity regularization term. The composite loss becomes:
\begin{equation}
\mathcal{L}_{\text{total}} = \mathcal{L}_{\text{task}} + \lambda \cdot \mathcal{L}_{\text{reg}},
\end{equation}
where:
\begin{equation}
\mathcal{L}_{\text{task}} = -\frac{1}{n} \sum_{i=1}^n \sum_{k=1}^K y_{ik} \log f_k(x_i),
\end{equation}
\begin{equation}
\mathcal{L}_{\text{reg}} = \frac{1}{|\mathcal{F}_{\text{spurious}}|} \sum_{j \in \mathcal{F}_{\text{spurious}}} \left\| \frac{\partial f(x)}{\partial x_j} \right\|^2.
\end{equation}

\subsubsection{Adversarial Training}
To reinforce robustness, we include adversarial examples generated using FGSM:
\begin{equation}
x^{\text{adv}} = x + \epsilon \cdot \text{sign} \left( \nabla_x \mathcal{L}_{\text{task}}(f(x), y) \right),
\end{equation}
and compute adversarial loss:
\begin{equation}
\mathcal{L}_{\text{adv}} = \mathbb{E}_{(x,y) \sim \mathcal{D}} \left[ \mathcal{L}_{\text{task}}(f(x^{\text{adv}}), y) \right].
\end{equation}
The final training objective integrates all three components:
\begin{equation}
\mathcal{L}_{\text{total}} = \mathcal{L}_{\text{task}} + \alpha \cdot \mathcal{L}_{\text{adv}} + \lambda \cdot \mathcal{L}_{\text{reg}}.
\end{equation}

\subsection{Iterative Refinement Procedure}

\begin{algorithm}[tbp]
\small  
\caption{\textit{XAI-Guided Model Refinement for Enhanced Adversarial Robustness}}
\label{alg:lime_refinement}
\begin{flushleft}
\textbf{INPUT:} Initial model $f_\theta$, training data $\mathcal{D}$, feature attribution model LIME, thresholds $(\tau, \epsilon, \delta)$, and hyperparameters $(\lambda, \alpha)$.\\
\textbf{OUTPUT:} Refined robust model $f_\theta^*$.\\
\textbf{Notations:} $\mathcal{F}_{\text{spurious}}$: Set of flagged spurious features, \\
$\mathcal{L}_{\text{task}}$: task-specific loss, $\mathcal{L}_{\text{adv}}$: adversarial loss, $\mathcal{L}_{\text{reg}}$: sensitivity regularization loss.
\end{flushleft}
\begin{algorithmic}[1]
\STATE Initialize model $f_\theta$ using standard training on $\mathcal{D}$
\REPEAT
    \STATE $A \leftarrow$ \textsc{FeatureAttribution}$(f_\theta, \mathcal{D}, \text{LIME})$
    \STATE $\mathcal{F}_{\text{spurious}} \leftarrow$ \textsc{IdentifySpuriousFeatures}$(A, f_\theta, \tau, \epsilon, \delta)$
    \STATE Apply refinement strategies:
    \begin{ALC@g}
        \STATE \textbullet~ \textsc{FeatureMasking}$(\mathcal{F}_{\text{spurious}}, \mathcal{D})$
        \STATE \textbullet~ \textsc{SensitivityRegularization}$(f_\theta, \mathcal{F}_{\text{spurious}}, \lambda)$
        \STATE \textbullet~ \textsc{AdversarialTraining}$(f_\theta, \mathcal{D}, \alpha)$
    \end{ALC@g}
    \STATE Re-train model $f_\theta$ with updated loss:
    \[
    \mathcal{L} = \mathcal{L}_{\text{task}} + \alpha \cdot \mathcal{L}_{\text{adv}} + \lambda \cdot \mathcal{L}_{\text{reg}}
    \]
\UNTIL{Convergence or robustness constraint is satisfied}
\STATE \textbf{return} $f_\theta$
\end{algorithmic}
\end{algorithm}

The training proceeds iteratively as shown in \textbf{Algorithm \ref{alg:lime_refinement}}. In each iteration, the model is updated using the composite loss, LIME attributions are recomputed, and spurious features are re-evaluated. The process converges when robustness metrics stabilize or when attribution maps exhibit consistent, class-relevant focus across perturbations. This iterative loop ensures that the model increasingly grounds its predictions in robust, interpretable features—bridging the divide between adversarial defense and attributional transparency.

\section{Theoretical Analysis: Bridging Interpretability and Robustness}

This section establishes a formal connection between interpretability and adversarial robustness through the lens of feature attribution, gradient sensitivity, and decision boundary smoothness. Drawing inspiration from robustness theory—particularly Lipschitz-based analyses such as CLEVER~\cite{weng2018evaluating}—we demonstrate how the proposed explainability-guided attention control framework reduces adversarial vulnerability by suppressing sensitivity to spurious features identified via LIME.

\subsection{Attribution-Constrained Model Behavior}

Let \( f: \mathbb{R}^d \rightarrow \mathbb{R}^K \) be a multi-class classifier mapping input vectors \( x \in \mathbb{R}^d \) to output logits \( f_1(x), \ldots, f_K(x) \). The predicted class is \( y = \arg\max_k f_k(x) \). An adversarial example is defined as a perturbed input \( x + \delta \) such that \( \arg\max_k f_k(x + \delta) \neq y \), with \( \|\delta\|_p \leq \epsilon \).

For differentiable models, the first-order Taylor expansion yields:
\begin{equation}
f_k(x + \delta) \approx f_k(x) + \langle \nabla_x f_k(x), \delta \rangle.
\end{equation}

The change in prediction can be bounded using the norm of the gradient \cite{zuhlke2025adversarial}\cite{fazlyab2023certified}:
\begin{equation}
\sup_{\|\delta\|_p \leq \epsilon} \left| f_y(x) - f_j(x + \delta) \right| 
\leq \epsilon \cdot \| \nabla_x (f_y(x) - f_j(x)) \|_q,
\end{equation}
where \( \|\cdot\|_q \) is the dual norm of \( \|\cdot\|_p \).

This sensitivity motivates the use of gradient regularization as a defense mechanism. However, instead of regularizing all features equally, we aim to regularize spurious features—those identified by LIME as irrelevant, unstable, or highly sensitive.

\subsection{Attribution-Aligned Gradient Suppression}

Let \( \beta_j(x) \) be the LIME attribution for feature \( x_j \). Define the normalized attribution vector:
\begin{equation}
a(x) = \frac{[\left| \beta_1(x) \right|, \ldots, \left| \beta_d(x) \right|]}{\sum_{j=1}^d |\beta_j(x)|}.
\end{equation}
The directional alignment between model gradients and attribution is given by:
\begin{equation}
\text{Align}(x) = \frac{ \langle \nabla_x f_y(x), a(x) \rangle }{ \|\nabla_x f_y(x)\|_2 \cdot \|a(x)\|_2 }.
\end{equation}
A low alignment implies that the model's gradient sensitivity is dominated by features not identified as important—an indicator of vulnerability to adversarial perturbations.

\subsection{Attribution-Aware Lower Bound on Adversarial Distortion}
We now formalize the relationship between attribution-driven feature selection and adversarial robustness by introducing an attribution-aware lower bound on the minimum adversarial distortion required to change the model's prediction. This bound draws on the concept of local Lipschitz continuity \cite{zuhlke2025adversarial}\cite{dong2025adversarially} while incorporating constraints derived from LIME-based feature attribution and sensitivity analysis.


Let \( f: \mathbb{R}^d \to \mathbb{R}^K \) be a classifier that is locally Lipschitz continuous with constant \( L_q \) in the \( \ell_p \) norm. Let \( x \in \mathbb{R}^d \) be an input sample with true label \( y = \arg\max_k f_k(x) \). Define \( \mathcal{F}_{\text{spurious}} \subset \{1, \ldots, d\} \) as the set of features identified via LIME as spurious—i.e., features exhibiting high attribution magnitude, high gradient sensitivity, or instability across perturbed neighborhoods.

Suppose that for all \( j \in \mathcal{F}_{\text{spurious}} \), the following conditions hold:
\begin{align}
|\beta_j(x)| &\leq \tau, \label{eq:irrelevance_bound} \\
\left| \frac{\partial f_y(x)}{\partial x_j} \right| &\leq \epsilon, \label{eq:gradient_bound}
\end{align}
where \( \tau \) and \( \epsilon \) are small, user-defined thresholds for attribution and sensitivity respectively.
Then, for any untargeted adversarial attack constrained within an \( \ell_p \) ball of radius \( \delta \), the minimum perturbation required to change the prediction satisfies:
\begin{equation}
\Delta_{\min}(x) \geq \min_{j \neq y} \frac{f_y(x) - f_j(x)}{L_q^{\text{eff}}(x)},
\label{eq:lower_bound}
\end{equation}
where the effective Lipschitz constant is:
\begin{equation}
L_q^{\text{eff}}(x) = \left\| \nabla_x f_y(x) \odot \left(1 - m_{\text{spurious}} \right) \right\|_q,
\end{equation}
and \( m_{\text{spurious}} \in \{0, 1\}^d \) is a binary mask such that:
\begin{equation}
m_{\text{spurious},j} =
\begin{cases}
1, & \text{if } j \in \mathcal{F}_{\text{spurious}}, \\
0, & \text{otherwise}.
\end{cases}
\end{equation}

Let us begin with the first-order Taylor expansion of the output logits:
\begin{equation}
f_k(x + \delta) \approx f_k(x) + \langle \nabla_x f_k(x), \delta \rangle.
\end{equation}
For an untargeted attack to succeed, there must exist some \( j \neq y \) such that:
\begin{equation}
f_j(x + \delta) \geq f_y(x + \delta).
\end{equation}
Using the above expansion, the difference in logits becomes:
\begin{equation}
f_y(x + \delta) - f_j(x + \delta) \approx f_y(x) - f_j(x) + \langle \nabla_x (f_y(x) - f_j(x)), \delta \rangle.
\end{equation}
To flip the prediction, it suffices to consider:
\begin{equation}
\langle \nabla_x (f_y(x) - f_j(x)), \delta \rangle \leq f_j(x) - f_y(x).
\end{equation}
Applying Hölder's inequality, we obtain:
\begin{equation}
|\langle \nabla_x (f_y(x) - f_j(x)), \delta \rangle| \leq \|\nabla_x (f_y(x) - f_j(x))\|_q \cdot \|\delta\|_p.
\end{equation}
Solving for the minimal norm of \( \delta \) that satisfies the inequality leads to:
\begin{equation}
\|\delta\|_p \geq \frac{f_y(x) - f_j(x)}{\|\nabla_x (f_y(x) - f_j(x))\|_q}.
\end{equation}

Now, assuming the spurious gradients are bounded by \( \epsilon \), and attribution scores satisfy the irrelevance constraint \( |\beta_j(x)| \leq \tau \), we suppress these components by zeroing them out in the gradient vector using the mask \( m_{\text{spurious}} \). Thus, we compute the effective Lipschitz constant:
\begin{equation}
L_q^{\text{eff}}(x) = \left\| \nabla_x f_y(x) \odot (1 - m_{\text{spurious}}) \right\|_q.
\end{equation}



This above analysis establishes that spurious features—those with high attribution scores but low semantic relevance or unstable importance—contribute to adversarial vulnerability by inflating the local gradient norm. By systematically identifying and suppressing such features via LIME-guided refinement (i.e., through masking and regularization), the model effectively reduces its sensitivity to adversarial perturbations. As a result, the effective Lipschitz constant \( L_q^{\text{eff}}(x) \) becomes smaller, yielding a higher lower bound on the minimum adversarial distortion \( \Delta_{\min}(x) \).

\section{Experiments and Discussions}

\subsection{Experiments Setup}
To evaluate the effectiveness of our proposed framework—\textit{Bridging Interpretability and Robustness through XAI-Guided Model Refinement}—we perform a comprehensive set of experiments across multiple datasets, model architectures, and adversarial threat models. The primary objective is to demonstrate that integrating LIME-based feature attribution into the model refinement loop enhances adversarial robustness and generalization without significantly degrading standard accuracy.

\subsubsection{Datasets}
We conduct experiments on three benchmark datasets that collectively provide a diverse evaluation setting: CIFAR-10, CIFAR-10-C and CIFAR-100.

   
   

\subsubsection{Model Architectures}

We adopt ResNet-18~\cite{he2016deep} as our base architecture due to its balance between performance and computational efficiency. Its skip connections enable stable training, while its widespread adoption ensures comparability with prior work in adversarial robustness.

\subsubsection{Adversarial Attack Configurations}

To evaluate the robustness of our models under adversarial conditions, we consider three distinct evaluation scenarios: two involving gradient-based adversarial attacks (FGSM and PGD) and one targeting robustness to natural distributional shifts.



In addition to adversarial perturbations, we also evaluate the model's robustness to natural distribution shifts using the CIFAR-10-C dataset. This dataset includes 19 types of common corruptions (e.g., noise, blur, weather, digital artifacts), each applied at five levels of severity. The goal of this evaluation is to assess the model’s out-of-distribution (OOD) generalization capacity and its ability to remain robust under realistic perturbations.

\subsection{Experimental Results}



\subsubsection{CIFAR-10 Dataset}



\begin{figure}[tbp]
   \centering
   \includegraphics[width=\linewidth]{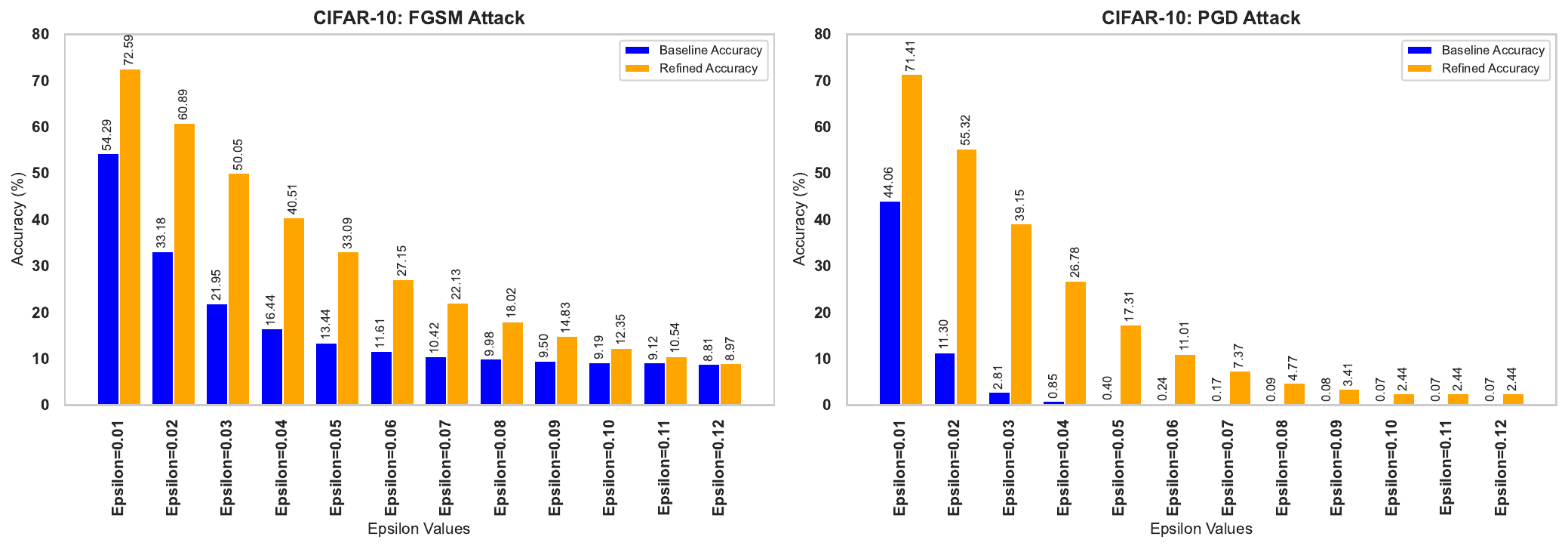}
   \caption{Robustness Analysis under FGSM and PGD Attacks on CIFAR-10. The refined model consistently demonstrates improved performance across different perturbation strengths.}
   \label{fig:robustness_analysis_c10}
\end{figure}

\begin{table*}[!t]
\centering
\begin{threeparttable}
\caption{Adversarial Robustness Analysis Against Common Corruptions on CIFAR-10-C}
\label{tab:robustness_analysis1}
\setlength{\tabcolsep}{3.5pt}    
\renewcommand{\arraystretch}{1.1} 
\small  
\begin{tabular}{@{}l|cccc|cccc|cccc@{}}
\toprule
\rowcolor{gray!10} & \multicolumn{4}{c|}{\textbf{$\epsilon = 0.01$}} & \multicolumn{4}{c|}{\textbf{$\epsilon = 0.02$}} & \multicolumn{4}{c}{\textbf{$\epsilon = 0.03$}} \\
\rowcolor{gray!10} & \multicolumn{2}{c}{FGSM} & \multicolumn{2}{c|}{PGD} & \multicolumn{2}{c}{FGSM} & \multicolumn{2}{c|}{PGD} & \multicolumn{2}{c}{FGSM} & \multicolumn{2}{c}{PGD} \\
\cmidrule{2-13}
\rowcolor{gray!10} \textbf{Corruption} & Base & Ref & Base & Ref & Base & Ref & Base & Ref & Base & Ref & Base & Ref \\
\midrule

\multicolumn{13}{l}{\cellcolor{gray!5}\textbf{\small Noise-Based}} \\
Gaussian & 79.06 & \textbf{90.69} & 65.87 & \textbf{90.32} & 43.4 & \cellcolor{blue!10}\textbf{88.54} & 13.68 & \cellcolor{blue!10}\textbf{86.57} & 25.17 & \cellcolor{blue!10}\textbf{86.13} & \cellcolor{red!10}3.29 & \cellcolor{blue!10}\textbf{71.44} \\
Shot & 79.58 & \textbf{90.63} & 65.97 & \textbf{90.32} & 43.39 & \cellcolor{blue!10}\textbf{88.16} & 14.25 & \cellcolor{blue!10}\textbf{85.84} & 24.49 & \cellcolor{blue!10}\textbf{85.61} & \cellcolor{red!10}2.75 & \cellcolor{blue!10}\textbf{68.18} \\
Impulse & 82.17 & \textbf{90.21} & 73.72 & \textbf{89.88} & 48.75 & \cellcolor{blue!10}\textbf{87.77} & 19.42 & \cellcolor{blue!10}\textbf{86.14} & 27.59 & \cellcolor{blue!10}\textbf{85.58} & \cellcolor{red!10}4.64 & \cellcolor{blue!10}\textbf{71.91} \\
Speckle & 80.53 & \textbf{90.69} & 67.96 & \textbf{90.38} & 45.28 & \cellcolor{blue!10}\textbf{88.13} & 15.08 & \cellcolor{blue!10}\textbf{86.44} & 26.05 & \cellcolor{blue!10}\textbf{85.94} & \cellcolor{red!10}3.24 & \cellcolor{blue!10}\textbf{71.21} \\

\midrule
\multicolumn{13}{l}{\cellcolor{gray!5}\textbf{\small Blur-Based}} \\
Defocus & 68.56 & \textbf{89.64} & 42.16 & \cellcolor{blue!10}\textbf{89.34} & 35.51 & \cellcolor{blue!10}\textbf{87.27} & \cellcolor{red!10}4.62 & \cellcolor{blue!10}\textbf{84.35} & 22.45 & \cellcolor{blue!10}\textbf{84.94} & \cellcolor{red!10}0.84 & \cellcolor{blue!10}\textbf{64.68} \\
Glass & 77.22 & \textbf{89.27} & 59.80 & \textbf{88.76} & 42.42 & \cellcolor{blue!10}\textbf{86.89} & 9.6 & \cellcolor{blue!10}\textbf{84.19} & 24.32 & \cellcolor{blue!10}\textbf{84.43} & \cellcolor{red!10}1.67 & \cellcolor{blue!10}\textbf{66.14} \\
Motion & 66.76 & \textbf{89.24} & 39.02 & \cellcolor{blue!10}\textbf{88.90} & 34.77 & \cellcolor{blue!10}\textbf{86.96} & \cellcolor{red!10}3.94 & \cellcolor{blue!10}\textbf{84.41} & 22.56 & \cellcolor{blue!10}\textbf{84.9} & \cellcolor{red!10}0.62 & \cellcolor{blue!10}\textbf{65.6} \\
Zoom & 67.40 & \textbf{90.03} & 41.50 & \cellcolor{blue!10}\textbf{89.60} & 34.36 & \cellcolor{blue!10}\textbf{87.39} & \cellcolor{red!10}3.98 & \cellcolor{blue!10}\textbf{84.49} & 21.78 & \cellcolor{blue!10}\textbf{85.06} & \cellcolor{red!10}0.62 & \cellcolor{blue!10}\textbf{64.2} \\
Gaussian & 66.27 & \textbf{88.10} & 37.28 & \cellcolor{blue!10}\textbf{87.27} & 33.68 & \cellcolor{blue!10}\textbf{84.22} & \cellcolor{red!10}3.51 & \cellcolor{blue!10}\textbf{76.63} & 22.35 & \cellcolor{blue!10}\textbf{79} & \cellcolor{red!10}0.54 & \cellcolor{blue!10}\textbf{53.06} \\

\midrule
\multicolumn{13}{l}{\cellcolor{gray!5}\textbf{\small Environmental}} \\
Snow & 78.18 & \textbf{90.17} & 63.03 & \textbf{89.78} & 43.39 & \cellcolor{blue!10}\textbf{87.92} & 12.12 & \cellcolor{blue!10}\textbf{86.04} & 25.66 & \cellcolor{blue!10}\textbf{85.82} & \cellcolor{red!10}2.26 & \cellcolor{blue!10}\textbf{70.16} \\
Frost & 69.37 & \textbf{89.41} & 44.39 & \cellcolor{blue!10}\textbf{88.80} & 31.68 & \cellcolor{blue!10}\textbf{86.99} & \cellcolor{red!10}4.8 & \cellcolor{blue!10}\textbf{82.19} & 17.23 & \cellcolor{blue!10}\textbf{84.49} & \cellcolor{red!10}1.04 & \cellcolor{blue!10}\textbf{56.66} \\
Fog & 28.49 & \cellcolor{blue!10}\textbf{57.69} & \cellcolor{red!10}4.98 & \cellcolor{blue!10}\textbf{34.31} & 10.64 & \cellcolor{blue!10}\textbf{57.08} & \cellcolor{red!10}0.09 & \cellcolor{blue!10}\textbf{7.42} & 6.84 & \cellcolor{blue!10}\textbf{57.48} & \cellcolor{red!10}0 & \cellcolor{blue!10}\textbf{2.01} \\

\midrule[\heavyrulewidth]
\rowcolor{gray!15}\textbf{Mean} & 70.30 & \textbf{87.15} & 50.47 & \textbf{84.81} & 37.27 & \textbf{84.78} & 8.76 & \textbf{77.89} & 22.21 & \textbf{82.45} & 1.79 & \textbf{60.44} \\
\rowcolor{gray!15}\textbf{Std} & ±13.87 & \textbf{±8.91} & ±18.55 & \textbf{±15.25} & ±9.60 & \textbf{±8.42} & ±5.78 & \textbf{±21.41} & ±5.29 & \textbf{±7.74} & ±1.37 & \textbf{±18.48} \\
\bottomrule
\end{tabular}
\begin{tablenotes}[flushleft]
\small
\item[$\dagger$] Bold numbers indicate best performance for each corruption type and attack
\item \cellcolor{blue!10}Blue highlighting indicates significant improvement ($>$50\%) over baseline
\item \cellcolor{red!10}Red highlighting indicates critical vulnerability ($<$5\% accuracy)
\end{tablenotes}
\end{threeparttable}
\end{table*}

\noindent Fig.~\ref{fig:robustness_analysis_c10} highlights significant robustness improvements achieved by the refined model. Under FGSM attacks, the refined model retains 72.59\% accuracy at $\epsilon=0.01$ (compared to 54.29\% for the baseline) and 50.05\% at $\epsilon=0.03$ (versus 21.95\%). Similarly, for PGD attacks, the refined model achieves 71.41\% accuracy at $\epsilon=0.01$, substantially outperforming the baseline's 44.06\%. 

\subsubsection{CIFAR-100 Dataset}



\begin{figure}[tbp]
   \centering
   \includegraphics[width=\linewidth]{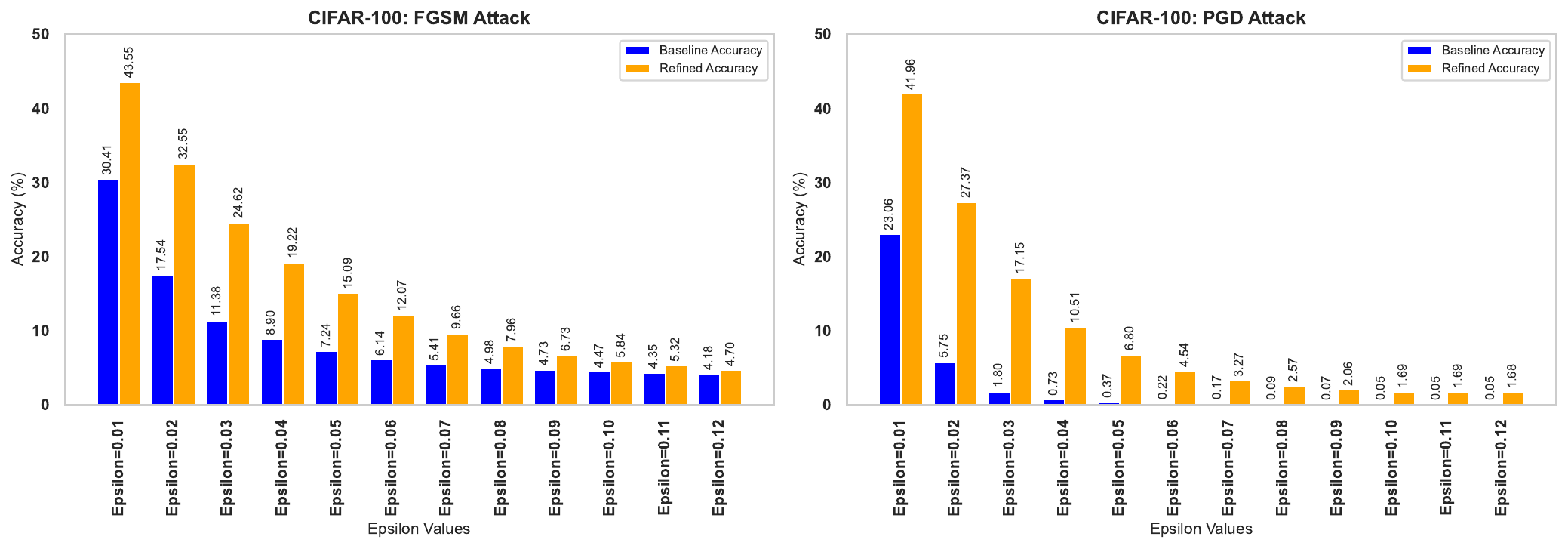}
   \caption{Robustness Analysis under FGSM and PGD Attacks on CIFAR-100. The refined model consistently achieves improved performance across varying perturbation strengths.}
   \label{fig:robustness_analysis_c100}
\end{figure}

\noindent As shown in Fig.~\ref{fig:robustness_analysis_c100}, the refined model demonstrates improved robustness on the more complex CIFAR-100 dataset. It achieves 43.55\% FGSM accuracy at $\epsilon=0.01$ (compared to 30.41\% for the baseline) and 24.62\% at $\epsilon=0.03$ (versus 11.38\%). For PGD attacks, the refined model achieves 41.96\% accuracy at $\epsilon=0.01$, significantly outperforming the baseline's 23.06\%. 


\subsubsection{CIFAR-10-C Dataset}




Table~\ref{tab:robustness_analysis1} further highlights the robustness of the refined model, particularly against noise-based corruptions. The refined model sustains $>$90\% accuracy under FGSM attacks at $\epsilon=0.01$ and $>$85\% under stronger PGD attacks. Cells highlighted in blue represent substantial gains ($>$50\% over baseline), while red highlights indicate vulnerabilities ($<$5\% accuracy). Across all corruption categories—noise, blur, and environmental—the refined model consistently achieves higher mean accuracy (87.15\% versus 70.30\% for FGSM and 84.81\% versus 50.47\% for PGD at $\epsilon=0.01$) with lower variability, indicating greater stability.

\section{Conclusion}

We proposed an attribution-guided training framework that integrates LIME explanations into the adversarial learning process to enhance model robustness and interpretability. By suppressing spurious and unstable features through masking, gradient regularization, and adversarial training, our method encourages reliance on semantically meaningful features.
Experiments on CIFAR-10, CIFAR-10-C, and CIFAR-100 demonstrate improved adversarial robustness and generalization with minimal loss in clean accuracy. Visualization and ablation studies confirm that our refinement approach leads to more stable and interpretable model behavior.
A theoretical analysis supports the connection between attribution alignment and adversarial resilience via an attribution-aware Lipschitz bound. Future work will explore alternative XAI methods, broader architectures, and applications in fairness-critical domains.
\balance 
\bibliographystyle{IEEEtran}
\bibliography{bibtext}

@inproceedings{krizhevsky2012imagenet,
  author = {A. Krizhevsky and I. Sutskever and G. E. Hinton},
  title = {ImageNet classification with deep convolutional neural networks},
  booktitle = {Advances in Neural Information Processing Systems (NeurIPS)},
  pages = {1097--1105},
  year = {2012}
}

@inproceedings{he2016deep,
  author = {K. He and X. Zhang and S. Ren and J. Sun},
  title = {Deep residual learning for image recognition},
  booktitle = {Proceedings of the IEEE Conference on Computer Vision and Pattern Recognition (CVPR)},
  pages = {770--778},
  year = {2016}
}

@article{samek2017explainable,
  author = {W. Samek and T. Wiegand and K.-R. Müller},
  title = {Explainable artificial intelligence: Understanding, visualizing and interpreting deep learning models},
  journal = {ITU Journal: ICT Discoveries},
  volume = {1},
  number = {1},
  pages = {39--48},
  year = {2017}
}

@inproceedings{szegedy2014intriguing,
  author = {C. Szegedy and W. Zaremba and I. Sutskever et al.},
  title = {Intriguing properties of neural networks},
  booktitle = {International Conference on Learning Representations (ICLR)},
  year = {2014}
}

@inproceedings{goodfellow2015explaining,
  author = {I. J. Goodfellow and J. Shlens and C. Szegedy},
  title = {Explaining and harnessing adversarial examples},
  booktitle = {International Conference on Learning Representations (ICLR)},
  year = {2015}
}

@inproceedings{madry2018towards,
  author = {A. Madry and A. Makelov and L. Schmidt and D. Tsipras and A. Vladu},
  title = {Towards deep learning models resistant to adversarial attacks},
  booktitle = {International Conference on Learning Representations (ICLR)},
  year = {2018}
}

@article{geirhos2019shortcut,
  author = {R. Geirhos and P. Rubisch and C. Michaelis et al.},
  title = {Shortcut learning in deep neural networks},
  journal = {Nature Machine Intelligence},
  volume = {2},
  pages = {665--673},
  year = {2019}
}

@inproceedings{ribeiro2016why,
  author = {M. T. Ribeiro and S. Singh and C. Guestrin},
  title = {Why should I trust you? Explaining the predictions of any classifier},
  booktitle = {Proceedings of the ACM SIGKDD International Conference on Knowledge Discovery and Data Mining (KDD)},
  year = {2016}
}

@article{wang2019representation,
  author = {L. Wang and Q. Liang},
  title = {Representation learning and nature encoded fusion for heterogeneous sensor networks},
  journal = {IEEE Access},
  volume = {7},
  pages = {39227--39235},
  year = {2019}
}

@inproceedings{lundberg2017unified,
  author = {S. M. Lundberg and S.-I. Lee},
  title = {A unified approach to interpreting model predictions},
  booktitle = {Advances in Neural Information Processing Systems (NeurIPS)},
  year = {2017}
}

@article{doshi2017rigorous,
  author = {F. Doshi-Velez and B. Kim},
  title = {Towards a rigorous science of interpretable machine learning},
  journal = {arXiv preprint arXiv:1702.08608},
  year = {2017}
}

@inproceedings{adebayo2018sanity,
  author = {J. Adebayo and J. Gilmer and M. Muelly et al.},
  title = {Sanity checks for saliency maps},
  booktitle = {Advances in Neural Information Processing Systems (NeurIPS)},
  year = {2018}
}

@inproceedings{slack2020fooling,
  author = {D. Slack and S. Hilgard and E. Jia et al.},
  title = {Fooling LIME and SHAP: Adversarial attacks on post hoc explanation methods},
  booktitle = {Proceedings of the AAAI/ACM Conference on AI, Ethics, and Society (AIES)},
  year = {2020}
}

@article{wang2024dense,
  author = {L. Wang and X. Li and Z. Zhang},
  title = {Dense Cross-Connected Ensemble Convolutional Neural Networks for Enhanced Model Robustness},
  journal = {arXiv preprint arXiv:2412.07022},
  year = {2024}
}

@article{wang2021explaining,
  author = {L. Wang and C. Wang and Y. Li and R. Wang},
  title = {Explaining the behavior of neuron activations in deep neural networks},
  journal = {Ad Hoc Networks},
  volume = {111},
  pages = {102346},
  publisher = {Elsevier},
  year = {2021}
}

@article{ross2018improving,
  author = {A. S. Ross and F. Doshi-Velez},
  title = {Improving the adversarial robustness and interpretability of deep neural networks by regularizing their input gradients},
  booktitle = {Proceedings of the AAAI Conference on Artificial Intelligence (AAAI)},
  year = {2018}
}

@article{dombrowski2019explanations,
  author = {A.-K. Dombrowski and M. Alber and C. J. Anders et al.},
  title = {Explanations can be manipulated and still be faithful: The case of feature attribution},
  booktitle = {International Conference on Learning Representations (ICLR)},
  year = {2019}
}

@inproceedings{zhou2022feature,
  title={Do feature attribution methods correctly attribute features?},
  author={Zhou, Yilun and Booth, Serena and Ribeiro, Marco Tulio and Shah, Julie},
  booktitle={Proceedings of the AAAI conference on artificial intelligence},
  volume={36},
  number={9},
  pages={9623--9633},
  year={2022}
}

@inproceedings{selvaraju2017grad,
  author = {R. R. Selvaraju and M. Cogswell and A. Das and R. Vedantam and D. Parikh and D. Batra},
  title = {Grad-CAM: Visual explanations from deep networks via gradient-based localization},
  booktitle = {Proceedings of the IEEE International Conference on Computer Vision (ICCV)},
  pages = {618--626},
  year = {2017}
}

@inproceedings{sundararajan2017axiomatic,
  author = {M. Sundararajan and A. Taly and Q. Yan},
  title = {Axiomatic attribution for deep networks},
  booktitle = {Proceedings of the 34th International Conference on Machine Learning (ICML)},
  pages = {3319--3328},
  year = {2017}
}

@inproceedings{wang2025explainability,
  title={Explainability-driven defense: Grad-cam-guided model refinement against adversarial threats},
  author={Wang, Longwei and Uddin, Ifrat Ikhtear and Qin, Xiao and Zhou, Yang and Santosh, KC},
  booktitle={Proceedings of the AAAI Symposium Series},
  volume={6},
  number={1},
  pages={49--57},
  year={2025}
}

@inproceedings{ranabhat2025multi,
  title={Multi-scale unrectified push-pull with channel attention for enhanced corruption robustness},
  author={Ranabhat, Robin Narsingh and Wang, Longwei and Qin, Xiao and Zhou, Yang and Santosh, KC},
  booktitle={Proceedings of the AAAI Symposium Series},
  volume={6},
  number={1},
  pages={34--41},
  year={2025}
}

@article{qin2024apbench,
  title={APBench: A unified availability poisoning attack and defenses benchmark},
  author={Qin, Tianrui and Gao, Xitong and Zhao, Juanjuan and Ye, Kejiang and Xu, Cheng-zhong},
  journal={Transactions on Machine Learning Research},
  year={2024}
}

@inproceedings{zou2024improving,
  title={Improving alignment and robustness with circuit breakers},
  author={Zou, Andy and Phan, Long and Wang, Justin and Duenas, Derek and Lin, Maxwell and Andriushchenko, Maksym and Kolter, J Zico and Fredrikson, Matt and Hendrycks, Dan},
  booktitle={The Thirty-eighth Annual Conference on Neural Information Processing Systems},
  year={2024}
}

@article{zhu2024neural,
  title={Neural polarizer: A lightweight and effective backdoor defense via purifying poisoned features},
  author={Zhu, Mingli and Wei, Shaokui and Zha, Hongyuan and Wu, Baoyuan},
  journal={Advances in Neural Information Processing Systems},
  volume={36},
  year={2024}
}

@article{weng2018evaluating,
  title={Evaluating the robustness of neural networks: An extreme value theory approach},
  author={Weng, Tsui-Wei and Zhang, Huan and Chen, Pin-Yu and Yi, Jinfeng and Su, Dong and Gao, Yupeng and Hsieh, Cho-Jui and Daniel, Luca},
  journal={arXiv preprint arXiv:1801.10578},
  year={2018}
}

@article{wang2024enhanced,
  title={Enhanced robustness by symmetry enforcement},
  author={Wang, Longwei and Ghimire, Aashish and Santosh, KC and Zhang, Zheng and Li, Xueqian},
  journal={IEEE CAI},
  year={2024}
}

@article{wang2024explainable,
  title={Explainable AI for 6G use cases: Technical aspects and research challenges},
  author={Wang, Shen and Qureshi, M Atif and Miralles-Pechu{\'a}n, Luis and Huynh-The, Thien and Gadekallu, Thippa Reddy and Liyanage, Madhusanka},
  journal={IEEE Open Journal of the Communications Society},
  year={2024},
  publisher={IEEE}
}

@article{chander2025toward,
  title={Toward trustworthy artificial intelligence (TAI) in the context of explainability and robustness},
  author={Chander, Bhanu and John, Chinju and Warrier, Lekha and Gopalakrishnan, Kumaravelan},
  journal={ACM Computing Surveys},
  volume={57},
  number={6},
  pages={1--49},
  year={2025},
  publisher={ACM New York, NY}
}

@article{seth2025bridging,
  title={Bridging the Gap in XAI-Why Reliable Metrics Matter for Explainability and Compliance},
  author={Seth, Pratinav and Sankarapu, Vinay Kumar},
  journal={arXiv preprint arXiv:2502.04695},
  year={2025}
}

@article{ennab2025advancing,
  title={Advancing AI Interpretability in Medical Imaging: A Comparative Analysis of Pixel-Level Interpretability and Grad-CAM Models},
  author={Ennab, Mohammad and Mcheick, Hamid},
  journal={Machine Learning and Knowledge Extraction},
  volume={7},
  number={1},
  pages={12},
  year={2025},
  publisher={MDPI}
}

@article{jagatheesaperumal2024enabling,
  title={Enabling trustworthy federated learning in industrial IoT: bridging the gap between interpretability and robustness},
  author={Jagatheesaperumal, Senthil Kumar and Rahouti, Mohamed and Alfatemi, Ali and Ghani, Nasir and Quy, Vu Khanh and Chehri, Abdellah},
  journal={IEEE Internet of Things Magazine},
  volume={7},
  number={5},
  pages={38--44},
  year={2024},
  publisher={IEEE}
}

@article{csahin2025unlocking,
  title={Unlocking the black box: an in-depth review on interpretability, explainability, and reliability in deep learning},
  author={{\c{S}}AHiN, Emrullah and Arslan, Naciye Nur and {\"O}zdemir, Durmu{\c{s}}},
  journal={Neural Computing and Applications},
  volume={37},
  number={2},
  pages={859--965},
  year={2025},
  publisher={Springer}
}

@article{zhou2025advancing,
  title={Advancing explainability of adversarial trained Convolutional Neural Networks for robust engineering applications},
  author={Zhou, Dehua and Song, Ziyu and Chen, Zicong and Huang, Xianting and Ji, Congming and Kumari, Saru and Chen, Chien-Ming and Kumar, Sachin},
  journal={Engineering Applications of Artificial Intelligence},
  volume={140},
  pages={109681},
  year={2025},
  publisher={Elsevier}
}

@article{zuhlke2025adversarial,
  title={Adversarial robustness of neural networks from the perspective of lipschitz calculus: A survey},
  author={Z{\"u}hlke, Monty-Maximilian and Kudenko, Daniel},
  journal={ACM Computing Surveys},
  volume={57},
  number={6},
  pages={1--41},
  year={2025},
  publisher={ACM New York, NY}
}

@article{dong2025adversarially,
  title={Adversarially robust neural architectures},
  author={Dong, Minjing and Li, Yanxi and Wang, Yunhe and Xu, Chang},
  journal={IEEE Transactions on Pattern Analysis and Machine Intelligence},
  year={2025},
  publisher={IEEE}
}

@article{fazlyab2023certified,
  title={Certified robustness via dynamic margin maximization and improved lipschitz regularization},
  author={Fazlyab, Mahyar and Entesari, Taha and Roy, Aniket and Chellappa, Rama},
  journal={Advances in Neural Information Processing Systems},
  volume={36},
  pages={34451--34464},
  year={2023}
}

@article{nayyem2024bridging,
  title={Bridging Interpretability and Robustness Using LIME-Guided Model Refinement},
  author={Nayyem, Navid and Rakin, Abdullah and Wang, Longwei},
  journal={arXiv preprint arXiv:2412.18952},
  year={2024}
}

@article{wall2025winsor,
  title={Winsor-cam: Human-tunable visual explanations from deep networks via layer-wise winsorization},
  author={Wall, Casey and Wang, Longwei and Rizk, Rodrigue and Santosh, KC},
  journal={arXiv preprint arXiv:2507.10846},
  year={2025}
}

@inproceedings{uddin2025expert,
  title={Expert-guided explainable few-shot learning for medical image diagnosis},
  author={Uddin, Ifrat Ikhtear and Wang, Longwei and Santosh, KC},
  booktitle={MICCAI Workshop on Data Engineering in Medical Imaging},
  pages={95--104},
  year={2025},
  organization={Springer Nature Switzerland Cham}
}

@article{wang2025bridging,
  title={Bridging Symmetry and Robustness: On the Role of Equivariance in Enhancing Adversarial Robustness},
  author={Wang, Longwei and Uddin, Ifrat Ikhtear and Santosh, KC and Zhang, Chaowei and Qin, Xiao and Zhou, Yang},
  journal={arXiv preprint arXiv:2510.16171},
  year={2025}
}

\end{document}